\documentclass{article} %
\usepackage{conf_arxiv,times}

\usepackage{amsmath,amsfonts,bm}

\def\eqref#1{equation~\ref{#1}}

\def\1{\bm{1}}

\DeclareMathAlphabet{\mathsfit}{\encodingdefault}{\sfdefault}{m}{sl}
\SetMathAlphabet{\mathsfit}{bold}{\encodingdefault}{\sfdefault}{bx}{n}

\usepackage[utf8]{inputenc} %
\usepackage[T1]{fontenc}    %
\usepackage{booktabs}       %
\usepackage{amsfonts}       %
\usepackage{nicefrac}       %
\usepackage{microtype}      %
\usepackage{graphicx}
\usepackage{natbib}
\usepackage{doi}
\usepackage{amssymb}
\usepackage{amsmath}
\usepackage{appendix}
\usepackage{algorithm}
\usepackage{algpseudocode}
\usepackage{import}
\usepackage{cleveref}
\usepackage{hyperref}
\usepackage{url}
\usepackage{fancyhdr}

\title{Adaptive Sparse Allocation with Mutual Choice \& Feature Choice Sparse Autoencoders}

\author{Kola Ayonrinde \\
        MATS \\
	\texttt{koayon@gmail.com} \\
}

\iclrfinalcopy %
\begin{document}

\fancyhead[L]{Preprint. Under review.}

\maketitle

\begin{abstract}
Sparse autoencoders (SAEs) are a promising approach to extracting features from neural networks,
enabling model interpretability as well as causal interventions on model internals.
SAEs generate sparse feature representations using a sparsifying activation function that implicitly defines
a set of token-feature matches.
We frame the token-feature matching as a resource allocation problem constrained by a total sparsity upper bound.
For example, TopK SAEs solve this allocation problem with the additional constraint that each token matches with at most $k$ features.
In TopK SAEs, the $k$ active features per token constraint is the same across tokens,
despite some tokens being more difficult to reconstruct than others.
To address this limitation, we propose two novel SAE variants, \emph{Feature Choice SAEs} and \emph{Mutual Choice SAEs},
which each allow for a variable number of active features per token.
Feature Choice SAEs solve the sparsity allocation problem under the additional constraint that each feature matches with at most $m$ tokens.
Mutual Choice SAEs solve the unrestricted allocation problem where the total sparsity budget can be allocated freely between tokens and features.
Additionally, we introduce a new auxiliary loss function, $\mathtt{aux\_zipf\_loss}$,
which generalises the $\mathtt{aux\_k\_loss}$ to mitigate dead and underutilised features.
Our methods result in SAEs with fewer dead features and improved reconstruction loss at equivalent sparsity levels
as a result of the inherent adaptive computation.
More accurate and scalable feature extraction methods provide a path towards
better understanding and more precise control of foundation models.
\end{abstract}

\section{Introduction}
\label{sec:intro}

Understanding the internal mechanisms of neural networks is a core challenge
in Mechanistic Interpretability.
Increased mechanistic understanding of foundation models could provide
model developers with tools to identify and debug undesirable model behaviour.

Dictionary learning with sparse autoencoders (SAEs) has recently
emerged as a promising approach for extracting sparse, meaningful, and
interpretable features from neural networks, particularly language
models \citep{eleuther_sae_2023, sharkey_sae_2022}.

One problem with wide SAEs for foundation models is that there are often many
dead features \citep{gated_saes_2024, templeton_scaling_monosemanticity_anthropic, gao_topk_openai}.
Dead features are features which are remain inactive across inputs, effectively
wasting model capacity and hampering efficient training.
Another problem is that approaches like TopK SAEs \citep{gao_topk_openai}
don't have a natural way to take advantage of Adaptive Computation: spending
more computation,  and crucially more features, to reconstruct more difficult tokens.

We frame the problem of generating sparse feature activations corresponding to
some given neural activations as a resource allocation problem, allocating the
scarce total sparsity budget between token-feature matches to maximise the
reconstruction accuracy.
Within this framing, we naturally
motivate two novel SAE variants which can provide Adaptive Computation:
Feature Choice SAEs and Mutual Choice SAEs (FC and MC SAEs respectively).
Feature Choice SAEs solve the sparsity
allocation problem under the additional constraint that each feature matches with at most m tokens.
Mutual Choice SAEs solve the unrestricted allocation problem where the total sparsity budget can
be allocated freely between tokens and features.
These approaches combine the Adaptive Computation of Standard SAEs with
the simple optimisation and improved performance of TopK SAEs.

\begin{figure}
    \centering
    \includegraphics[width=0.8\linewidth]{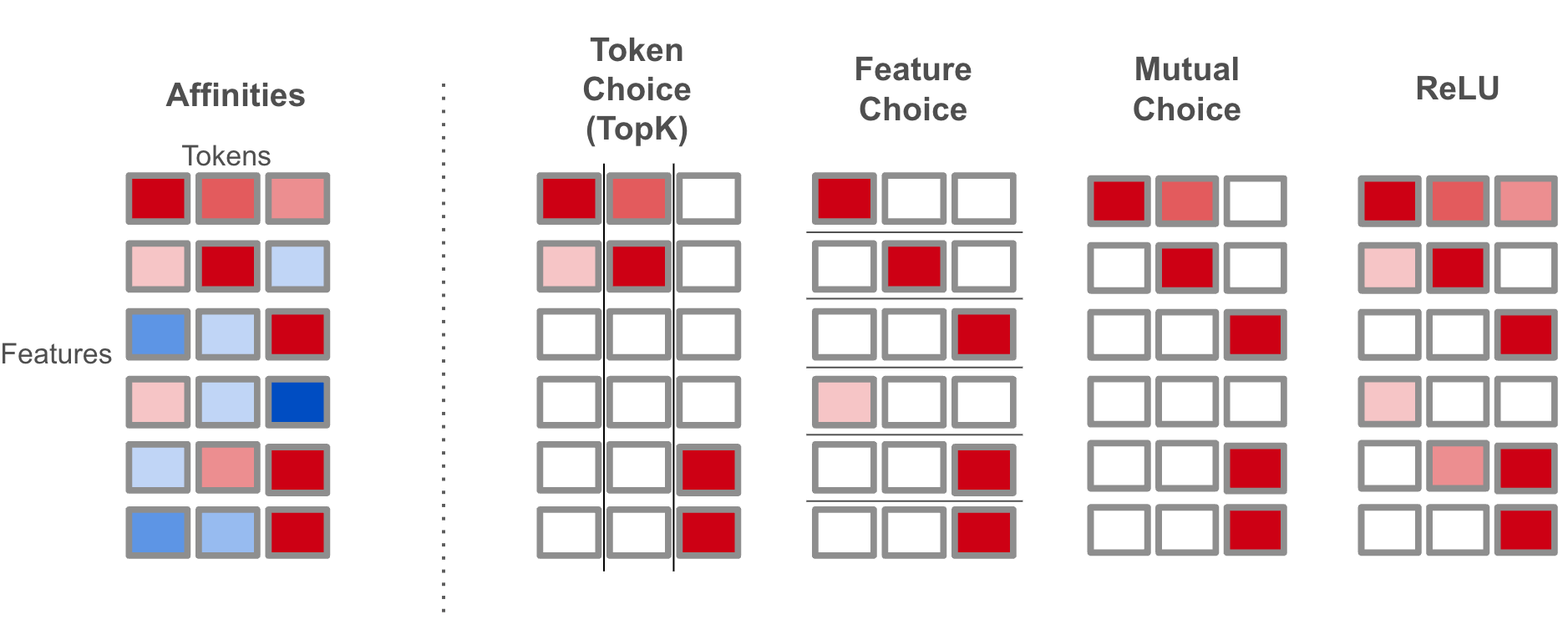}
    \caption{
    A comparison of the pre-activation affinities and the resulting
    feature activations following different sparsifying activation functions.
    Red and blue represent positive and negative affinities respectively,
    with deeper colours representing larger magnitudes.
    In the first three approaches we have a total sparsity budget of 6.
    \textit{Affinities (Far-left)}:
    The token-feature affinities $\textbf{Z'}$,
    before any sparsifying activation function.
    \textit{Token Choice/TopK (Center-left)}:
    We activate the top $k$ features corresponding to each token.
    Note that there are features that don't fire in this batch,
    which could lead to dead features.
    \textit{Feature Choice (Center)}:
    For each feature, it activates corresponding to the top $m$ tokens with the highest affinity.
    Note that all features fire in this batch.
    \textit{Mutual Choice (Center-right)}:
    The elements with the largest magnitude affinities activate,
    regardless of their token or feature affiliations.
    \textit{ReLU/Standard (Far-right)}:
    All strictly positive elements activate.
    Here we allow low-magnitude feature activations which may be false positives and which cause the $L_0$ to be higher.
    Jump-ReLU SAEs can be seen as a special case of ReLU SAEs
    where the activation threshold is non-zero.
    }
    \label{fig:sparse_affinity_matrices}
\end{figure}

Our contributions are as follows:

\begin{itemize}
\item
  We provide a framing for sparsifying activation functions in SAEs
  as a solution to a resource allocation problem.
\item
  We introduce two new SAE architectures: the \textbf{Mutual Choice
  SAE} and \textbf{Feature Choice SAE}, which are Pareto improvements
  on both standard SAEs and TopK SAEs. Additionally,
  the Feature Choice approach is, to our knowledge, the first SAE training method
  which reliably results in \textit{~zero dead features} even at large scale.
\item
  We show that our methods naturally enable \textit{Adaptive Computation}: using
  more features to reconstruct more difficult tokens. Instead of setting
  the number of features per token as a fixed $k$, we fix $\mathbb{E}[k]$ as a hyperparameter
  and allow the model to learn how to allocate the sparsity budget, without
  increasing computational overhead.
\item
    We propose a novel auxiliary loss function, $\mathtt{aux\_zipf\_loss}$, which
    mitigates under-utilisation of features and better utilises the SAE's full capacity.

\end{itemize}

We believe that with increasingly accurate approaches to feature extraction,
it will become possible to connect sparse features over many layers and understand
how models are computing outputs in a mechanistic, circuits-driven fashion.
In particular, given that Feature Choice SAEs generally have no dead features,
they can scale reliably to very large autoencoders, which are
likely to be necessary for effective reconstruction on large
foundation models such as GPT-4 \citep{openai_gpt-4_2024} or Llama 3 \citep{llama3model}.

The remainder of this paper is organized as follows: In \textit{Section 2}, we review the sparse autoencoder literature and the challenges in training SAEs.
\textit{Section 3} details our resource allocation framing for the sparsifying activation function and introduces the Mutual Choice and Feature Choice SAE architectures (MC and FC respectively).
In \textit{Sections 4 \& 5}, we describe our methods and experimental setup and in \textit{Section 6} we present results from training our sparse autoencoders.
We discuss our findings and the limitations of our work in \textit{Section 7} and conclude in \textit{Section 8} with directions for future research.

\section{Related Work}
\label{sec:related_work}

\hypertarget{adaptive_computation}{%
\subsection{Adaptive Computation}\label{adaptive_computation}}

In \textbf{Adaptive Computation}, neural networks decide how much compute (and/or which parameters) to apply to a given
input example
\citep{graves2017adaptivecomputationtimerecurrent, xue2023adaptivecomputationelasticinput}.
Ideally, the model should learn to apply less compute to easier examples and
more compute to more difficult examples in order to maximise
performance within a compute budget.
In our setting, we consider the token-feature matches to be the scarce quantity to allocate,
where we say that a token matches with a feature if the
feature is activated on that token.

\subsection{Sparse Autoencoders}
\label{sparse-autoencoders}

\textbf{Sparse Autoencoders (SAEs)} \citep{lee_sparse_deep_belief_net_2007, unsupervised_learning_sae_2012, sparse_coding_image_modelling_2014} learn an over-complete basis, or dictionary, of sparsely activating features.
The feature activations, $z$, correspond to their associated neural activations, $x$, via the feature dictionary. In particular, we can write an SAE as:
\begin{align}
z &= \sigma_s(\text{Enc}(x)) \in \mathbb{R}^F \\
\hat{x} &= \text{Dec}(z) \in \mathbb{R}^N
\end{align}

where $\sigma_s$ is a sparsifying activation function (e.g. ReLU), $\text{Dec}$ is an affine map and $x \in \mathbb{R}^N$
\footnote{In practice, there may be additional pre-processing and post-processing of the neural activations, $x$.}.

SAEs are trained to minimize the Reconstruction Error (Mean Squared Error) between $x$ and $\hat{x}$.
This reconstruction error term is combined with an optional
Sparsity Loss term (for example, an $L_1$ penalty to induce sparsity)
and an optional Auxiliary Loss term to reduce dead features:
\begin{equation}
    \mathcal{L}(x) = |x - \hat{x}|_2^2 + \lambda_1 \mathcal{L}_\text{sparsity}(z) +  \lambda_2 \mathcal{L}_\text{aux}(x, z, \hat{x})
\end{equation}

\cite{gated_saes_2024, templeton_scaling_monosemanticity_anthropic, gao_topk_openai}
have shown that decomposing neural activations using the SAE feature dictionary
allows for increased human interpretability of models even at
model sizes comparable to frontier foundation models.

\hypertarget{topk-saes}{%
\subsection{TopK SAEs}\label{topk-saes}}

\textbf{TopK SAEs} \citep{gao_topk_openai} use a TopK activation function instead of the $L_1$
penalty to induce sparsity, as in \cite{k-sparse_autoencoders_2014}.
Though in the standard $L_1$ SAE formulation, the number of features-per-token is variable,
in the TopK formulation the features-per-token is fixed at the same $k$ for all tokens.
We hypothesize that having a fixed $k$ is a key drawback of the TopK method.
Variable $k$ values introduce Adaptive Computation which can focus
more of the token-feature matching budget on more difficult tokens.

In concurrent work, \cite{bussmann_batchtopk_2024} introduce BatchTopK
which is closely analogous to our Mutual Choice SAEs and also provides adaptive computation.
However, they do not deal with the problem of underutilised features.

\hypertarget{dead_features}{%
\subsection{Dead Features}\label{dead_features}}

SAE features which remain inactive across many inputs are known as \textbf{dead features}.
\cite{anthropic_sae_towards_monosemanticity_bricken} declare a feature to be dead
when it hasn't fired for at least 1e7 tokens.
Dead features present a challenge especially when scaling to
larger models and wider autoencoders.
For example, \cite{templeton_scaling_monosemanticity_anthropic} find 64.7\% of
features are dead for their autoencoders with 34M features.
Our Feature Choice approach naturally results in \textit{~zero dead features} by ensuring that each feature activates for every batch.

\hypertarget{aux_k}{%
\subsection{Auxiliary k loss function}\label{aux_k}}

\cite{gao_topk_openai} propose the auxiliary loss function, $\mathtt{aux\_k\_loss}$
to reduce the proportion of dead features.
Given the SAE residual $e = x - \hat{x}$, they define
the auxiliary loss $\mathcal{L}_\text{aux}$ = $|e - \hat{e}|^2$,
where $\hat{e}$ = Dec($z_\text{dead}$) is the reconstruction using the top $k_\text{aux}$ dead features.
\cite{gao_topk_openai} find fewer dead features when using the $\mathtt{aux\_k\_loss}$ for SAE training.

However, the $\mathtt{aux\_k\_loss}$ is only applied to features which qualify as dead.
We apply a similar auxiliary loss, the $\mathtt{aux\_zipf\_loss}$, to underutilised, but not yet dead, features.

\section{Background}
\label{sec:background}

\subsection{Sparsifying Activation Functions as Resource Allocators}
\label{sec:allocation}

We consider the following many-to-many matching problem:

\begin{itemize}
\item
  We have $F$ features and a batch of $B$ tokens.
  We would like to have at most $M$ token-feature matches,
  where we say that a token matches with a feature if the feature is activated on that token
  \footnote{Here we follow the Mixture of Expert literature in abusing notation slightly to refer to neural activations corresponding to a given token position as a "token".}
  .
\item
  We would like to allocate our budget of $M$ token-feature matches such that the reconstruction error is minimised.
\end{itemize}

Formally, we seek a reconstruction-error optimal weighted subgraph
$\mathcal{H} \subseteq \mathcal{G} = \{\{1,...,B\} \times \{1, ..., F\}, \textbf{E}\}$
where $\mathcal{H}$ has at most $M$ edges, for $M \ll BF$.
The edge weights can be viewed as the token-feature affinities:
the pre-sparsifying activation feature magnitudes $\textbf{z'}$.
\footnote{As an illustrative example of the same problem,
we can imagine a university which is able to confer at most $M$ degrees
where a student may take many degree subjects and
a degree subject may admit many students.}

This problem doesn't immediately admit an efficient solution because it is currently
unspecified how the edge weights contribute to the token reconstruction error.
We make a simplifying assumption that we denote the \textbf{Monotonic Importance Heuristic} -
the edges with the largest edge weights are likely to represent the most important
contributions to the reconstruction error.
With this heuristic, we can solve the problem of allocating token-feature matches
by choosing the \textit{M} edges with the
largest magnitude edge weights as the edges for our subgraph $\mathcal{H}$.

\begin{figure}
    \centering
    \includegraphics[width=0.8\linewidth]{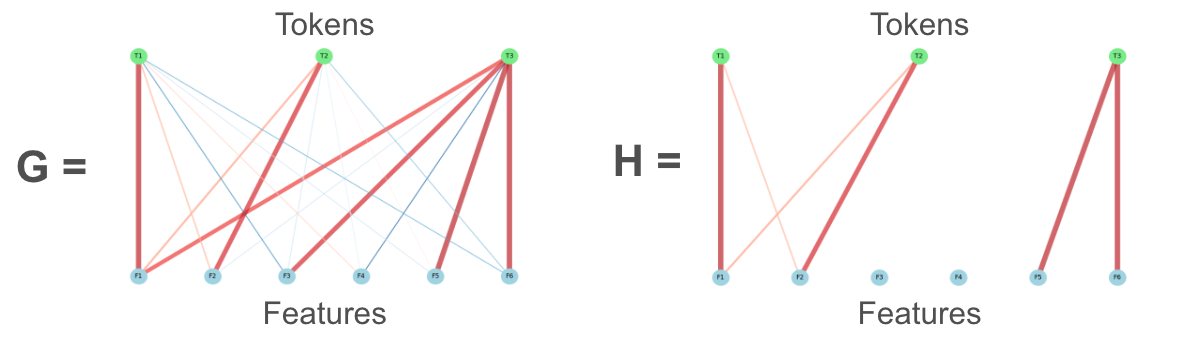}
    \caption{$\mathcal{G}$ (left) is a weighted bipartite graph $\mathcal{G} = \{\{T_1, T_2, T_3\} \times \{F_1, F_2, ..., F_6 \}, \textbf{E}\}$. Edge weights represent pre-activation affinities with red and blue representing positive and negative values respectively. We are seeking a subgraph $\mathcal{H} \subseteq \mathcal{G}$ with $M=6$ edges.
    Here we have defined $\mathcal{H}$ (right) by the TopK method for $k=2$; we select the 2 edges from each token with the largest edge weights.}
    \label{fig:subgraph}
\end{figure}

We can equivalently view this allocation problem as choosing a binary mask
$\textbf{S} \in \mathbb{R}^{B\times F}$ with at most \textit{M} non-zero elements
which maximises reconstruction accuracy.
This mask is to be element-wise multiplied with a token-feature affinity matrix $ \textbf{Z'} \in \mathbb{R}^{B\times F}$.
Applying the Monotonic Importance Heuristic, we are looking for the mask S such that
$\sum_{i,j}\textbf{Z}_{i,j} = \sum_{i,j}\textbf{Z'}_{i,j} \odot \textbf{S}_{i,j} $ is maximised.

\textbf{TopK SAEs:}
We can now see the TopK SAE approach as a special case of the above allocation problem,
with the additional constraint that the number of features per token
is at most $k$ for each token.
In other words, $\sum_i(\textbf{S}_{i,j}) = k$ $\forall j$, where $M = kB$.
This leads to the solution of $\textbf{S} = \text{TopKIndices}(\textbf{Z'}, \text{dim} = -1) $,
i.e. $\textbf{S}$ picks out the $k$ features with the highest affinity for each token.
Here $\sigma_s(\textbf{z'}) = \textbf{S} \odot \textbf{z'}$;
element-wise multiplication with $\textbf{S}$ defines our
sparsifying activation function $\sigma_s$.

We now consider two other variants of this problem displayed in \cref{fig:sparse_affinity_matrices}:

\textbf{Feature Choice SAEs:}
Whilst TopK SAEs require each token to match with at most $k$ features,
we instead add the constraint that each feature matches with at most $m$ tokens.
This is equivalently a constraint on the columns of S rather than its rows:
$\sum_j(\textbf{S}_{i,j}) = m$ $\forall i$, where $M = mF$.
This leads to the solution of $\textbf{S} = \text{TopKIndices}(\textbf{Z'}, \text{dim} = -1)$,
i.e. $\textbf{S}$ picks out the $m$ tokens with the highest affinity for each feature.

\textbf{Mutual Choice SAEs:}
Here we don't add any additional constraints and allow any choice of token-feature matching.
This leads to the solution of $\textbf{S} = \text{TopKIndices}(\textbf{Z'}, \text{dim} = (0,1))$,
i.e. \textbf{S} picks out the largest elements of the \textbf{Z'} affinity matrix, regardless of their position.

The Feature Choice (FC) and Mutual Choice (MC) sparsifying activation functions
can be seen as having the desirable properties of the TopK activation
(for example, preventing activation shrinkage,
reducing the impact of noisy, low magnitude activations,
allowing for a progressive recovery code,
enabling simple model comparison,
not requiring sparsity losses which are in conflict with the reconstruction loss etc.)
but whilst allowing for Adaptive Computation, see \cref{fig:sparse_affinity_matrices}.
Since we do not have constraints on the number of features per token in
FC or MC, it's possible for one token to activate (i.e. match with)
more features than another token.

\subsubsection{Analogy To Mixture of Experts Routing}

We choose our naming here to align with the study of token-expert matching in the Mixture of Experts paradigm, where there is a close analogy.
Token Choice MoE routing strategies \citep{shazeer_moe_2017, switch_transformer_fedus} have the constraint that each token can be
routed to at most $k$ experts allowing for expert imbalances.
On the other hand, Expert Choice routing strategies \citep{expertchoice_zhou} have the
constraint that each expert processes at most m tokens,
which eliminates the possibility of underutilised experts
and allows tokens to be routed to a variable number of experts.

Here the intuitions are ``\textit{each token picks
the $k$ experts to be routed to}'' and ``\textit{each expert picks
the $m$ tokens to be routed to that expert}'' for Token Choice and Expert Choice respectively.
The variety of MoE routing algorithms is explored in \cite{vision_routers_unified}.
The \textit{Feature Choice} approach we propose is directly analogous to the \textit{Expert Choice} approach in MoEs
and \textit{TopK} SAEs are directly analogous to \textit{Token Choice} MoEs. For this reason, we will call TopK SAEs, Token Choice SAEs to unify our notation.

\subsection{Distribution of Feature Densities}
\label{sec:feature_dist}

\begin{figure}
    \centering
    \includegraphics[width=0.6\linewidth]{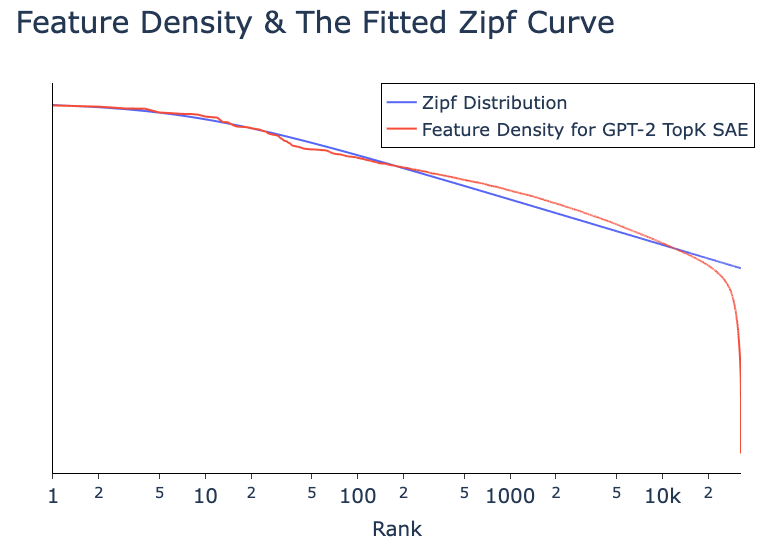}
    \caption{The Feature Density Curve fits a Zipf curve with $R^2=0.982$. The middle part of the feature density distribution (feature 100-20,000) fit the Zipf curve with $R^2$=0.996}
    \label{fig:zipf_r_squared}
\end{figure}

When analysing the distribution of feature densities in open-source SAEs from
\cite{gao_topk_openai},
we find that the distributions typically follow a power law
described by the Zipf distribution with $R^2$ = 0.982, see \cref{fig:zipf_r_squared}.

We note that although the Zipf distribution well fits most of the distribution,
there is a considerable residual at the lower end of the distribution (the 20k+ rank tokens).
We suggest that these features are underutilised.
Underutilised features see fewer gradient updates than other features
leading to a self-reinforcing cycle
of these features being less useful and hence further underutilised until they die.

We will refer to these underutilised features as \textbf{dying features}.
We define a dying feature as a feature which is one of the 25\% least prevalent features
which is also <60\% as prevalent as we might predict from the prevalence rank
of the feature using the fitted Zipf curve.

Previous approaches to dealing with dead features either resampled dead features
\citep{anthropic_sae_towards_monosemanticity_bricken}
or applied gradients to dead features
(for example $\mathtt{aux\_k\_loss}$ \citep{gao_topk_openai})
but they didn't address dying features.
We hypothesise that many of the revived dead features were still not appropriately utilised.

To address the problem of dying features,
we add an additional auxiliary loss for dying features,
which is a natural generalisation of the $\mathtt{aux\_k_loss}$.
Given the SAE residual $e = x - \hat{x}$, we define
the auxiliary loss $\mathcal{L}_\text{aux\_zipf} = |e - \hat{e}|^2$,
where $\hat{e}$ = Dec($z_\text{dying}$) is the reconstruction
using the top $k_\text{aux}$ dying features.
We can think of this $\mathtt{aux\_zipf\_loss}$ acting \textit{preventatively} on features which could be at risk of
becoming dead and acting \textit{rehabilitatively} on features which have been recently revived.
In this way, we reduce the proportion of both dead and dying features.

\hypertarget{feature_choice_zipf}{%
\subsection{Choosing the Feature Choice Constraint}\label{feature_choice_zipf}}

In the Feature Choice approach, there remains the question of how to distribute the sparse feature
activations across the feature dimension.

The simplest approach to this is to take $m_i = \nicefrac{M}{F}$ for
all \textit{i}, where \textit{F} is the number of features.
In other words, each feature can pick exactly \textit{m} tokens to
process. We call this approach \textit{Uniform Feature Choice}.

Uniform prevalence is a natural way to organize the features
so that a feature firing provides maximal information about the token,
under the assumption that all features provide approximately equal information.
However, we have seen that in existing open-source SAEs,
all features are not equivalently prevalent.
Instead, they are approximately Zipf-law distributed.
To maintain this distribution of feature density
we choose $m_i \sim$ Zipf($\alpha$, $\beta$),
where Zipf represents a truncated Zipf distribution
and \textit{i} is the rank of a given feature in terms of feature density.
\begin{equation}
    m_{i} = \text{Zipf}(i) \propto \frac{1}{(i + \beta)^{\alpha}}
\end{equation}

We call the Feature Choice approach where the $m_i$ are Zipf-distributed, \textit{Zipf Feature Choice}, henceforth simply \textit{Feature Choice}.
\section{Methods}
\label{sec:methods}
Our approach is as follows:

\begin{itemize}
\item
  Given the Zipf exponent and bias hyperparameters, $\alpha$ and $\beta$,
  \footnote{We may obtain these hyperparameters by performing a hyperparameter sweep.
  Alternatively, if we have trained SAE of the same dimensions, we may run inference with this SAE over a large dataset (100 million tokens)
  and track the number of times each feature activates
  as its \textit{feature density}
  and each feature's relative feature density \textit{rank}.
  We can then fit the (rank, feature density) pairs to a Zipf distribution
  and estimate the exponent and bias parameters, $\alpha$ and $\beta$.
  For GPT-2 sized residual stream activations (n=768),
  we find $\alpha \approx 1.0, \beta \approx 6.8$.
  In our experiments, we fix the exponent to exactly 1 for simplicity
  and we find that we may reuse similar $\alpha$ and $\beta$ parameters
  across SAE widths.
  }
  , we use \cref{alg:zipf_m} to determine the estimated feature density
  for each ranked feature.
  We use the estimated feature densities to define the threshold
  for dying features for the $\mathtt{aux\_zipf\_loss}$.

\item
  We then train Mutual Choice SAEs with both the $\mathtt{aux\_zipf\_loss}$ and $\mathtt{aux\_k\_loss}$.

\item
  Finally, we optionally fine-tune these SAEs with the Feature Choice activation function,
  adding the constraint on the number of tokens that each feature should process.
  Here there are no auxiliary loss terms.
\end{itemize}

The sparsifying activation functions $\sigma_s$ for each approach are all TopK activations where the
TopK is taken over the feature dimension for Token Choice (i.e. TopK),
the batch dimension for Feature Choice and
all dimensions for Mutual Choice.
\hypertarget{setup}{%
\section{Experimental Setup}\label{setup}}

\textbf{Inputs:} We train our sparse autoencoders on the layer 6 residual stream activations of GPT-2 small \citep{gpt2_paper}.
We use a context length of 64 tokens for all experiments.
We preprocess the activations by subtracting the mean over the $d_\text{model}$
dimension and normalize all inputs to unit $L_2$ norm.
We shuffle the activations for training our SAEs (as in \cite{nanda_open_2023}).
All experiments are performed without feature resampling, unless otherwise specified.

\textbf{Hyperparameters:}
We tune learning rates based on \cite{gao_topk_openai}
suggestion that the learning rate scales like $\sqrt{n}$.
We use the AdamW optimizer \citep{adam_optim_2017} and a batch size of 1,536
\footnote{For the Feature Choice approach, it's important to have sufficiently
large minibatch sizes so that each feature is expected to activate every few minibatches. }.
We train each SAE for 10,000 steps or until convergence.
We use a weight decay of 1e-5 and apply gradient clipping.
We analyse SAE widths from 4x to 32x the width of the neural activations with 0.8\% feature sparsity.
We use gradient accumulation for larger batch sizes.
We do not perform extensive hyperparameter sweeps.

\textbf{Evaluation:} After training, we evaluate autoencoders on sparsity $L_0$, reconstruction (MSE) and the difference on the model's final (Cross-Entropy) loss.
We report a standard normalized version of the loss recovered (\%).
We additionally evaluate our SAEs' interpretability using \cite{eleuther_autointerp_open_2024}'s automated interpretability (AutoInterp) process.
We report the percentage of dead features across models.

\textbf{Baselines:} We compare our SAEs against Standard (ReLU) SAEs and TopK SAEs
\footnote{We do not explicitly test against Gated SAEs, but \cite{gao_topk_openai}
find that TopK SAEs perform similarly or better than Gated SAEs with $1.5\times$ less compute to convergence.
}.
\section{Results}
\label{sec:results}

\begin{figure}[h]
    \centering
    \includegraphics[width=0.75\linewidth]{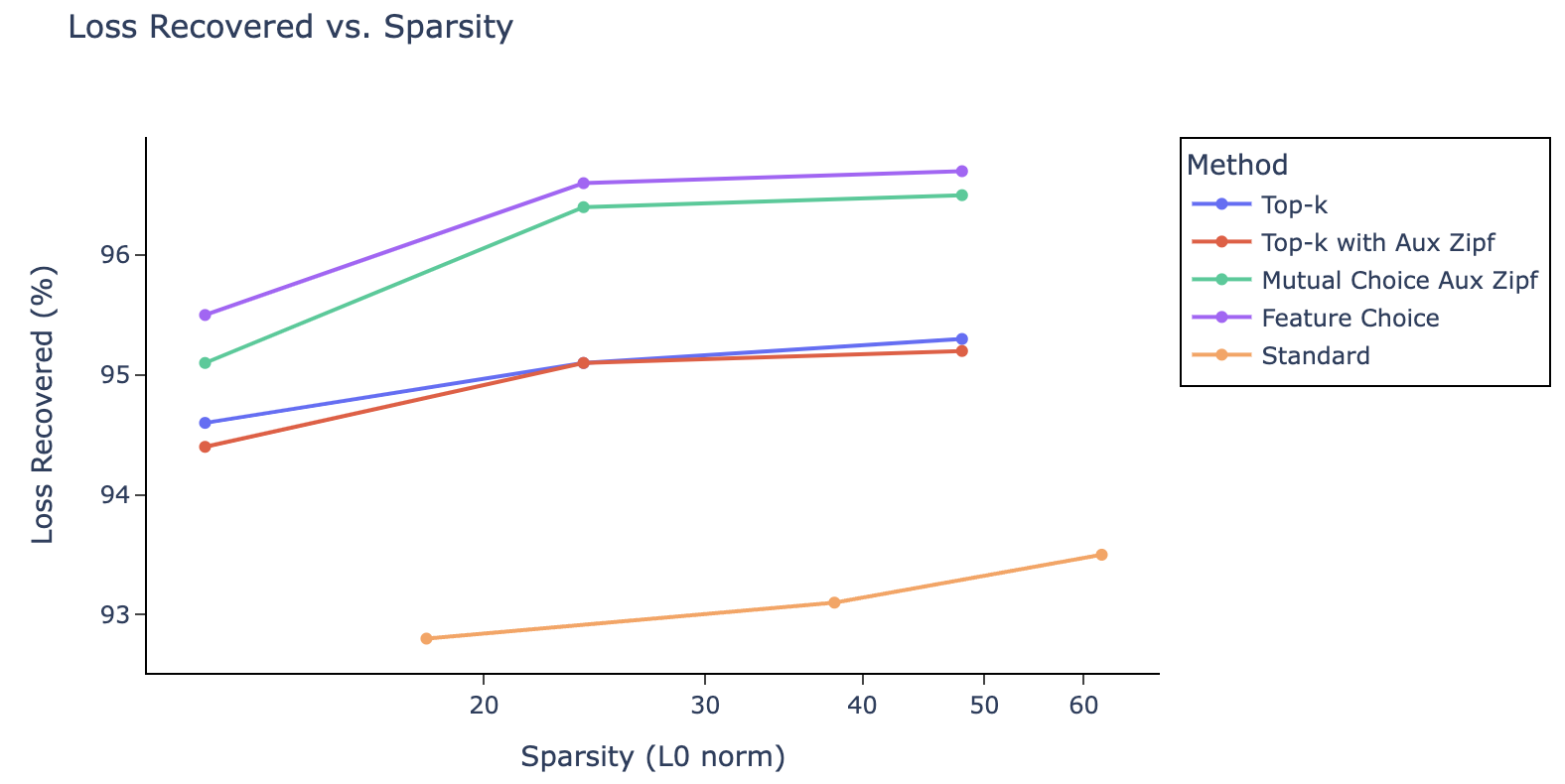}
    \caption{SAEs trained with the Mutual Choice activation function, and those finetuned with the Feature Choice activation function have better downstream loss recovered at equivalent sparsity levels.}
    \label{fig:main_pareto}
\end{figure}

We find that Feature Choice SAEs are a Pareto improvement upon the Token Choice TopK SAEs,
as in \cref{fig:main_pareto}.

\begin{figure}[h]
    \centering
    \includegraphics[width=0.75\linewidth]{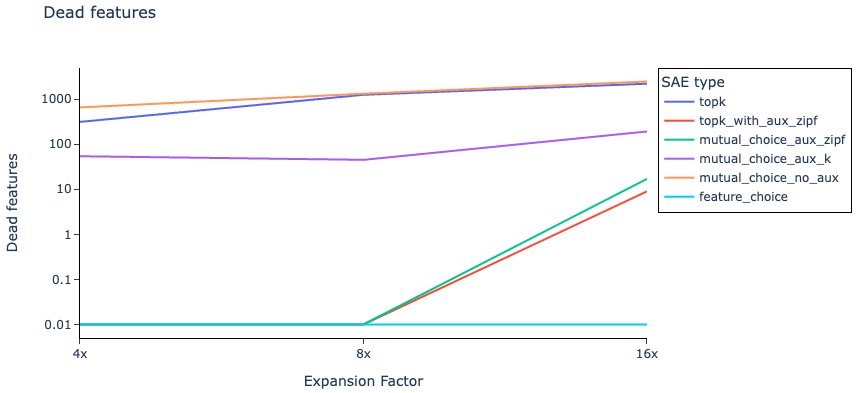}
    \caption{The Mutual Choice SAEs with the $\mathtt{aux\_zipf\_loss}$
    applied and the Feature Choice SAEs have fewer dead features than SAEs trained without the $\mathtt{aux\_zipf\_loss}$.}
    \label{fig:main_dead_dying_features}
\end{figure}

We find that both Mutual Choice and Feature Choice SAEs provide better utilisation of the SAE
capacity with fewer dead features
than comparable SAE methods \cref{fig:main_dead_dying_features}.

\begin{table}[h]
\centering
\begin{tabular}{|l|c|c|c|}
\hline
\textbf{SAE Type} & \textbf{6k latent dim} & \textbf{34m dim} & \textbf{16m dim} \\
\hline
Standard SAE \citep{anthropic_sae_towards_monosemanticity_bricken} & 9.0\% & - & >90.0\% \\
\hline
SAE++ (w/ resampling) \citep{templeton_scaling_monosemanticity_anthropic} & 5.1 \% & 64.7\% & - \\
\hline
Token Choice (TopK) SAE \citep{gao_topk_openai} & 0.0\% & - & 7.0\% \\
\hline
Feature Choice SAE (Ours) & \textbf{0\%} & \textbf{0\%} & \textbf{0\%} \\
\hline
\end{tabular}
\caption{Percentage of Dead Features for Different SAE Types and Model Sizes}
\label{table:speculative_dead_features}
\end{table}

\cite{templeton_scaling_monosemanticity_anthropic}'s 34 million latent SAEs have a dead feature rate of 64.7\% (with resampling);
\cite{gao_topk_openai}'s 16 million latent SAEs have a dead feature rate of 90\% without mitigations and 7\% with mitigations.
We find that Feature Choice SAEs can be trained with a 0\% dead feature rate for both 16m and 34m latent SAEs as in \cref{table:speculative_dead_features}.

\hypertarget{discussion}{%
\section{Discussion}\label{discussion}}

We summarize the comparison between our approach and related approaches
in \cref{tab:sae_comparison}:

\begin{table}[h]
\centering
\begin{tabular}{|l|l|l|l|l|}
\hline
\text{SAE Type} & \text{Performance} & \text{Dead Features} & \text{Auxiliary Losses Required} \\
\hline
\text{Standard \citep{anthropic_sae_towards_monosemanticity_bricken}} & \text{Weak} & \text{Many} & \text{$L_1$ Sparsity loss} \\
\hline
\text{SAE ++ \citep{templeton_scaling_monosemanticity_anthropic}} & \text{Better} & \text{Many} & \text{Scaled $L_1$ sparsity loss} \\
\hline
\text{Jump-ReLU \citep{jumprelu_sae_deepmind}} & \text{Better} & \text{Fewer} & \text{$L_1$ Sparsity loss} \\
\hline
\text{Token Choice (TopK) \citep{gao_topk_openai}} & \text{Better} & \text{Fewer} & $\mathtt{aux\_k}$  \\
\hline
\text{Mutual Choice (ours)} & \textbf{Best} & \textbf{Fewer} & $\mathtt{aux\_k}$ and $\mathtt{aux\_zipf}$  \\
\hline
\text{Feature Choice (ours)} & \textbf{Best} & \textbf{None} & \textbf{None} \\
\hline
\end{tabular}
\caption{Comparison of SAE Types}
\label{tab:sae_comparison}
\end{table}

To our knowledge, we present the first SAE training process which
explicitly contains two separate phases:
first \emph{Mutual Choice training}, then \emph{Feature Choice training}.
We speculate that there may be performance benefits to phased training.

We believe that conceiving of the role of SAE encoders as defining matching/routing
algorithms (similar to other Adaptive Computation work, for
example, within the Mixture of Experts literature) could be a valuable
intuition pump for further improvements to SAE architectures.

Our approaches can also be combined with the MDL-SAE \citep{ayonrinde_interpretability_2024}
formulation which treats \textbf{conciseness} (\textbf{Description Length})
as the relevant quantity for evaluation and model selection rather than sparsity ($L_0$).
The Description Length of a set of feature activations is a function of the sparsity,
the distribution of activation patterns for each feature and the SAE width.

\hypertarget{zipf-top-m}{%
\subsection{Zipf Distributed Features}\label{zipf-top-m}}

We find that constraining the number of tokens per feature with the Zipf distribution
outperforms using the uniform distribution by >10\% model loss recovered. The large drop in performance
using the uniform distribution compared to the Zipf distribution gives additional evidence
that naturally occurring features are not uniformly distributed.

We hypothesise that the reason that features appear to be Zipf distributed may be strongly analogous to the
reason that the frequency of words in natural languages like English are also Zipf distributed.
Words are the semantic units of sentences.
Since features are the semantic units of computation within language models,
a similar mechanism could explain the empirical tendency for the densities of SAE features
to tend towards the Zipf distribution.

In the Computational Linguistics literature, it is well known that the distribution of words
in natural languages approximately follows a Zipf distribution \citep{orig_zipf}.
For example, in written English text, the empirical distribution over words
(treated as a categorical variable) can be modelled as Zipf($\alpha$, $\beta$) = Zipf(1, 2.7).

We speculate that the Preferential Attachment Theory explanation
\citep{zipf_least_effort, zipf_hierarchical_scaling}
for the tendency of words to be Zipf-distributed,
which states that frequently used words (features) tend to be used more often,
may be analogously applicable here.
At initialisation, there is some variance in feature prevalence.
The tokens which are initially most highly activated early in training
receive the most gradient signal and are most refined,
leading to a virtuous cycle where they are more effective and useful
for a larger part of the feature density spectrum.
\cref{fig:feature_dist_at_init} illustrates this dynamic over time.
We would be excited about further work detailing an explicit mechanism for
why features in Neural Networks tend towards being Zipf distributed.

We also note that, for humans, some concepts appear more prevalent in the world than others.
Hence, if features do map to human-interpretable concepts,
we might also expect variance in the prevalence of SAE features.

\hypertarget{token_difficulty}{%
\subsection{Adaptive Computation for Varying Token Difficulty}\label{token_difficulty}}

One benefit of the Mutual Choice and Feature Choice approaches is that
they allow for \textit{Adaptive Computation} - difficult to reconstruct tokens
(which may represent complex or rare concepts) can be reconstructed using more features,
whereas other tokens which are more straightforward can use fewer features.
Where Token Choice (TopK) SAEs suggest that all tokens are equal;
MC and FC SAEs suggest that, in fact, some tokens are more equal
than others \cite{singer_all_animals}.

As an extreme example, we might expect that the activations resulting from the $\mathtt{<BOS>}$
token are relatively easy to reconstruct, considering they are both very common and
have exactly the same value every time. We might expect that an effective SAE could learn to
productively reallocate the sparsity budget that would have been spent on the $\mathtt{<BOS>}$
token to more difficult tokens, thus increasing the SAE's effective capacity.
We provide an example of the features per token distribution in \cref{features_per_token_appendix}.

\hypertarget{dead_features_disc}{%
\subsection{Dead Features}\label{dead_features_disc}}

Our methods, especially Feature Choice SAEs, have many fewer dead features than other approaches
(typically zero) without complex and compute-intensive resampling procedures \citep{anthropic_sae_towards_monosemanticity_bricken}.
This is especially important for large SAEs where the problem of dead features is typically more
significant.
We hypothesize that this might be an additional reason for the improved performance: all features
receive some gradient signal at every step.

We note that this simplified approach eliminates the need for complex
resampling procedures. This approach suggests a
simplified procedure for SAE training and model selection than previous
SAEs with fixed sparsity budget.

\hypertarget{philosophical_notes}{%
\subsection{Philosophical Notes}\label{philosophical_notes}}

In \ref{zipf-top-m}, we discussed the fact that features which map to
human-understandable concepts and are learned from human-generated data
(like internet text) are likely approximately Zipf distributed.
We would be interested in how this varies for data which is less obviously
human-generated, such as images or video.
We speculate that an information theoretic analysis of the features
of the natural world which have predictive power could
provide evidence for ontological theories.
We suggest that exploring the features apparent in the natural world
information-theoretically could be a valuable lens for
theorists working in the Naming \& Necessity paradigm \citep{kripke}
or within Natural Abstractions Theory \citep{natural_abstractions}.

\hypertarget{limitations}{%
\subsection{Limitations}\label{limitations}}

\textbf{Appeals to the Monotonic Importance Heuristic:}
In \cref{sec:allocation} we defined the Monotonic Importance Heuristic (MIH) -
the assumption that the importance of a feature is monotonically increasing in feature activation magnitude.
We use this assumption when we choose the feature activations with the largest magnitudes with our TopK-style
activation functions.
TopK SAEs also implicitly assume the MIH (Monotonic Importance Heuristic).
We can think of Jump-ReLU SAEs and Gated SAEs as relaxing this assumption slightly
to a weak MIH.
For Jump-ReLU and Gated SAEs, the importance of activations is still related
to their magnitude and they still filter out any low magnitude feature activations;
however, the filtering threshold varies for each feature.
So Jump-ReLU SAEs do not have to make magnitude comparisons
across features which may have different natural scales.
It may be, however, that there are low magnitude activations which,
within a certain context, are nonetheless critically important in capturing
information which is useful for reconstruction and/or downstream model performance.
These important but low magnitude activations are difficult to
capture with our current SAE approaches
\footnote{Standard ReLU SAEs do allow low magnitude feature activations but at the expense of
failing to filter out noisy low magnitude activations, which can be seen as false positives
\citep{jumprelu_sae_deepmind}. }.

Though the weight-sharing form of Gated SAEs \citep{gated_saes_2024} implicitly encodes the weak MIH prior,
the non-sharing form does not.
Weight-sharing Gated SAEs, however, tend to perform better.
The improved performance of approaches which encode the MIH prior could
be considered as evidence for the truth of the claim.
Alternatively, we might note that the Monotonic Importance Heuristic
acts as an inductive bias for our models.
Good inductive biases often allow models to perform better at first;
however, with increased scale we may not need such inductive biases
and may prefer allowing the model to learn more of the solution independently
\citep{xiao2024rethinkingconventionalwisdommachine}, \citep{bitterlesson}.

\textbf{Generalisation Across Modalities:}
We tested our SAEs within the domain of language.
We currently don't know to what extent our results generalise across modalities,
especially to inherently continuous modalities like audio and images.
We would be excited about future work applying similar techniques to
Interpretability problems in a wider range of modalities.

\textbf{Evaluation:}
The Mechanistic Interpretability field doesn't currently have widely agreed upon metrics
for evaluating Sparse Autoencoders.
Disentanglement benchmarks like \cite{huang2024ravelevaluatinginterpretabilitymethods} have been proposed as well as evaluation on tasks
where the ground truth is known \citep{adamboardgamesSAEs}.
Developing a more comprehensive suite of benchmarks for SAEs would
help us to have higher confidence in our comparisons between SAE variants.
\hypertarget{conclusion}{%
\section{Conclusion}\label{conclusion}}
We introduce the Feature Choice and Mutual Choice SAEs as a simple drop-in change
to the sparsifying activation function in Sparse Autoencoders. We also provide a
new auxiliary loss, $\mathtt{aux\_zipf\_loss}$, which prevents dying features and hence allows the SAE to
more fully utilize all of its features without wasting capacity.

We would be excited about future work that defines scaling laws for
Features Choice SAE performance on even larger models than the ones we use in this
work.
We believe a promising direction for future work is dissecting the
phenomenology of features {\citep{anthropic_sae_towards_monosemanticity_bricken}} conditioning on
both feature prevalence and feature activation magnitude.
We might ask
the question ``what kinds of features appear very often in naturally occurring
datasets and what does this imply about the predictive elements of the natural world?''.
We believe this direction could prove fruitful for both
Machine Learning researchers and philosophers of science.

\hypertarget{acknowledgments}{%
\subsection*{Acknowledgments}\label{acknowledgments}}
Thanks to Callum McDougall, Matthew Wearden, Jacob Drori, Alex Cloud, Joseph Miller and Evžen Wybitul for comments on early drafts.

Thanks to Lee Sharkey, Andrew Gritsevskiy, Derik Kauffman, Hans Gundlach, and Jan
Kulveit for useful conversations. Thanks to Sandy Tanwisuth for advice and encouragement.

Thanks to MATS for providing the space to do this work and to EleutherAI for providing compute.

\bibliography{references}
\bibliographystyle{conf_arxiv}

\newpage
\appendix %

\section{Feature Density Moves Towards A Zipf Distribution Throughout Training}

\begin{figure}[h]
    \centering
    \includegraphics[width=0.75\linewidth]{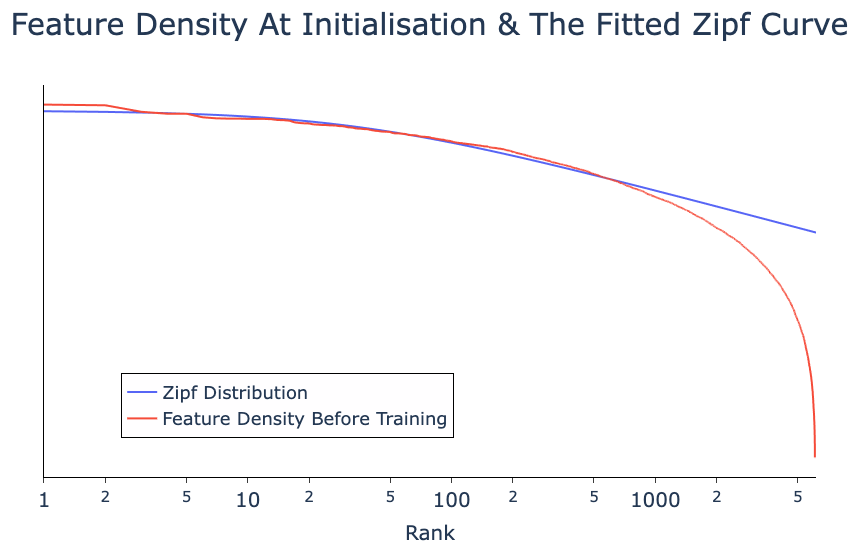}
    \caption{At initialisation, the feature distribution diverges from the Zipf distribution significantly at a smaller feature rank than after training the feature dictionary.}
    \label{fig:feature_dist_at_init}
\end{figure}

We can compare this with \cref{fig:zipf_r_squared} where the drop-off comes much later, at 24k vs at 3k.

\section{Inference with Feature Choice and Mutual Choice SAEs}

To run inference on our SAE variants, you can do batch inference with the method exactly as in the training setup.
However to do single token (or single sequence) it may be beneficial to instead impute a threshold value and swap out the activation function to use this value instead with a JumpReLU style approach \cite{jumprelu_2019}.

\section{Neural Feature Matrix Loss}

One problem with SAEs is undesirable feature splitting.
Feature splitting occurs when an SAE finds a sparse combination of existing directions that allows for a smaller $L_0$ \citep{ayonrinde_interpretability_2024}. For example, \cite{anthropic_sae_towards_monosemanticity_bricken} note that a model may learn dozens of features which all represent the letter "P" in different contexts in order to maintain low sparsity.

In order to reduce feature splitting we propose two additional auxiliary losses: $\mathtt{nfm\_loss}$ and $\mathtt{nfm\_inf\_loss}$.

The Neural Feature Matrix (NFM) is defined as $\mathtt{nfm}(W) = \hat{W}\hat{W}^T$ for a weight matrix $W$.
The NFM is a symmetric square matrix which describes the correlation of different rows in the matrix $W$.

We define $\mathtt{nfm\_loss} = |\mathtt{nfm}(W_{Dec})|_F$ and $\mathtt{nfm\_inf\_loss} = \frac{1}{F}\sum_i{max({\mathtt{nfm}(W_{Dec})}_i})$.

For our largest runs we apply both of these auxiliary losses with small weight.
Empirically we find this seems to reduce undesirable feature splitting and avoid a failure mode we call "dictionary collapse" when many features of the decoder dictionary start to align with each other.

\hypertarget{features_per_token_appendix}{%
\section{Adaptive Computation Allows For A Variable Number of Features Per Token}\label{features_per_token_appendix}}

\begin{figure}[h]
    \centering
    \includegraphics[width=0.75\linewidth]{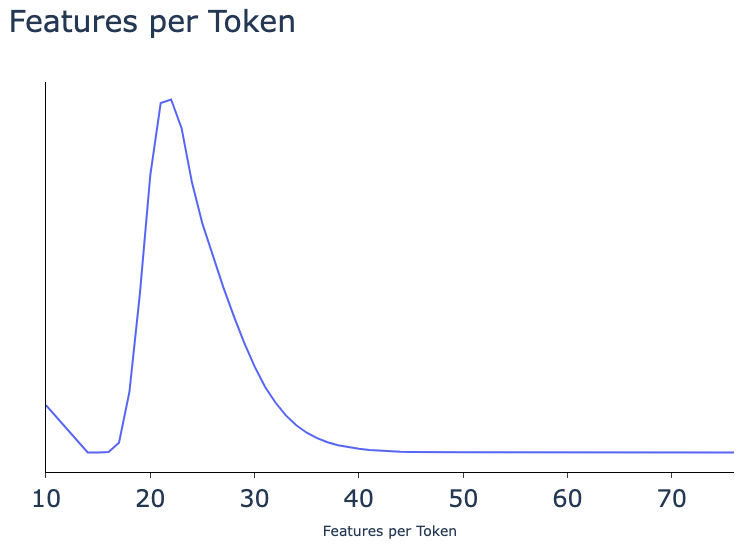}
    \caption{The number of features per token is variable as a result of the Adaptive Computation.}
    \label{fig:enter-label}
\end{figure}

We note that there is a bimodal distribution of features per token. The SAE allocates close to the main mode of features to each token, attempting to allocate more features to harder tokens and less features to easier tokens. The $\mathtt{<BOS>}$ token is responsible for the lower peak - this token is much easier to reconstruct and so it is prudent for the network to allocate fewer features that token. In the ideal SAE, the $\mathtt{<BOS>}$ token could be reconstructed with a single feature, rather than the 10 features required here. Future work might look into optimisations that would allow the $\mathtt{<BOS>}$ token to be single-feature reconstructed as a test case of allowing even greater variance in features per token.

\hypertarget{progressive_codes}{%
\section{Progressive Codes}\label{progressive_codes}}

\cite{gao_topk_openai} describe learning a progressive code using their Multi TopK
loss. In their setting, the Multi TopK loss (a weighted sum of TopK losses for different values
of $k$) is required because the SAE generally "overfits" to the value of $k$ which harms a progressive code.
In our case, the SAE is robust to having variable $k$ values even for the same token depending on the context of the batch.
Empirically, we obtain progressive codes for a greater range of values of $k$ than in the TopK case.

\hypertarget{determine_feat_dist}{%
\section{Determining The Feature Distribution}\label{determine_feat_dist}}

Given the Zipf exponent and bias hyperparameters, $\alpha$ and $\beta$, we use Algorithm 1 \cref{alg:zipf_m} to determine the estimated feature density
  for each ranked feature.
  We use the estimated feature densities to define the threshold
  for dying features for the $\mathtt{aux\_zipf\_loss}$.

\begin{algorithm}

\caption{Calculate Zipf Feature Distribution}
\label{alg:zipf_m}
\begin{algorithmic}[1]
\Require $k$, $F$, $B$, $\beta$, $\alpha$, $m_\text{max}$
\Ensure $m$: array of size $F$
\State num\_interactions $\gets B \times k$
\State zipf\_sum $\gets \sum_{i=1}^{F} \frac{1}{(i + \beta)^{\alpha}}$
\State $N_\text{approx} \gets \frac{\text{num\_interactions}}{\text{zipf\_sum}}$
\For{$i = 1$ \textbf{to} $F$}
    \State $m_i \gets \min\left(\left\lfloor\frac{N_\text{approx}}{(i + \beta)^{\alpha}}\right\rfloor,\ m_\text{max}\right)$
\EndFor
\State \Return $m$
\end{algorithmic}

\end{algorithm}

\hypertarget{mih}{%
\section{Monotonic Importance Heuristic}\label{mih}}

In \cref{sec:background} we appeal to the Monotonic Importance Heuristic (MIH), as in TopK SAEs, in order to simplify our allocation problem. Empirically we find that this works well though we discuss the case for not using the MIH in \cref{limitations}.

One theoretical (though informal) motivation for the MIH is as follows.
Since the decoder dictionary is fixed to unit norm,
the norm of any feature's contribution to the output is exactly the
magnitude of the feature activation to which it corresponds.
Hence if there's limited cancellation between features (which is likely in an N dimensional space where the per-token sparsity is much less than N) then we might expect the component of $\hat{x}$ in any feature direction to be very close to the feature activation for that feature.
In particular, consider a feature with a small magnitude of $\varepsilon$.
This feature can only possibly influence the reconstruction loss
by at most $\varepsilon |e|$ where $e = x - \hat{x}$.
Features corresponding to larger activations can plausibly influence the reconstruction loss by a greater amount.

\end{document}